\documentclass[letterpaper]{article} 
\usepackage{aaai25}  
\usepackage{times}  
\usepackage{helvet}  
\usepackage{courier}  
\usepackage[hyphens]{url}  
\usepackage{graphicx} 
\usepackage{natbib}  
\usepackage{caption} 
\usepackage{algorithm}
\usepackage{algorithmic}

\usepackage{xcolor}
\definecolor{mygreen}{RGB}{112,173,71}
\usepackage{newfloat}
\usepackage{listings}
\DeclareCaptionStyle{ruled}{labelfont=normalfont,labelsep=colon,strut=off} 
\floatstyle{ruled}
\newfloat{listing}{tb}{lst}{}
\floatname{listing}{Listing}
\usepackage{booktabs}
\usepackage{tabularx}
\usepackage{makecell}
\usepackage{multicol,multirow}
\usepackage{tcolorbox}
\usepackage{amsmath}
\usepackage{amsfonts}

\newcommand{\blue}[1]{\textcolor{blue}{#1}}
\newcommand{\green}[1]{\textcolor{mygreen}{#1}}
\newcommand{\red}[1]{\textcolor{red}{#1}}

\title{Cobra: Extending Mamba to Multi-Modal Large Language Model for Efficient Inference}
\author{
    Han Zhao\textsuperscript{\rm 1 2}\equalcontrib,
    Min Zhang\textsuperscript{\rm 1}\equalcontrib,
    Wei Zhao\textsuperscript{\rm 2},
    Pengxiang Ding\textsuperscript{\rm 2},
    Siteng Huang\textsuperscript{\rm 3},
    Donglin Wang\textsuperscript{\rm 2}\thanks{Corresponding author.}
}
\affiliations{
    \textsuperscript{\rm 1}Zhejiang University
    \textsuperscript{\rm 2}Westlake University
    \textsuperscript{\rm 3}DAMO Academy, Alibaba Group\\
    zhaohan34@westlake.edu.cn, zhangmin@westlake.edu.cn, zhaowei@westlake.edu.cn, \\
    dingpengxiang@westlake.edu.cn, siteng.huang@gmail.com, wangdonglin@westlake.edu.cn
}

\usepackage{bibentry}

\begin{document}

\maketitle

\begin{abstract}
In recent years, applying multi-modal large language models (MLLMs) in various fields has achieved remarkable success. However, as the foundation model for many downstream tasks, MLLMs comprise the well-known Transformer network, which has a less efficient quadratic computation complexity. In this paper, we introduce Cobra, a multi-modal large-scale language model built upon a state-space model, which has demonstrated significant potential in efficiently handling long sequences with fast inference and linear scalability concerning sequence length. Specifically, Cobra involves replacing Transformer-based backbone models (\textit{e.g.}, LLaMA or Phi) with pre-trained Mamba language models. We then empirically explore effective strategies for aligning visual and textual modalities and integrating various pre-trained Mamba model variants with visual encoders. Experiments across various multi-modal benchmarks demonstrate that: (i) Cobra performs $3 \times \sim 4 \times$ faster than the most computationally efficient state-of-the-art methods, \textit{e.g.}, LLaVA-Phi and MobileVLM v2. Additionally, its performance is significantly enhanced thanks to the implementation of linear sequential modeling. (ii) Cobra fine-tunes a small parameter ($\sim$ 48\% of model parameters), leading to a significant improvement in overall performance compared to LLaVA. The project page is available at:  https://sites.google.com/view/cobravlm.

\end{abstract}

\section{Introduction}
\label{sec:intro}

Multi-modal large language models (MLLMs) have recently achieved impressive results across a variety of downstream tasks, including multi-modal content generation~\cite{lu2022unifiedio, wu2024nextgpt}, vision-based question answering~\cite{openai2023gpt4v, gao2023llamaadapter, liu2023llava1.5, liu2023llava}, and embodied intelligence~\cite{brohan2023rt2, kim2024openvla, ding2024quarvla}. By effectively aligning pre-trained large language models with visual modalities, MLLMs have demonstrated a strong ability to comprehend and navigate complex visual-language contexts. These advancements not only highlight the versatility of MLLMs but also pave the way for further research into more nuanced and sophisticated applications. For example, as MLLMs continue to evolve, there is potential for significant improvements in areas such as real-time interaction in dynamic environments, cross-modal retrieval tasks, and the seamless integration of language and visual processing in everyday technology. 

\begin{figure*}[htbp]
  \centering
  \includegraphics[width=0.90\textwidth, height=0.50\textwidth]{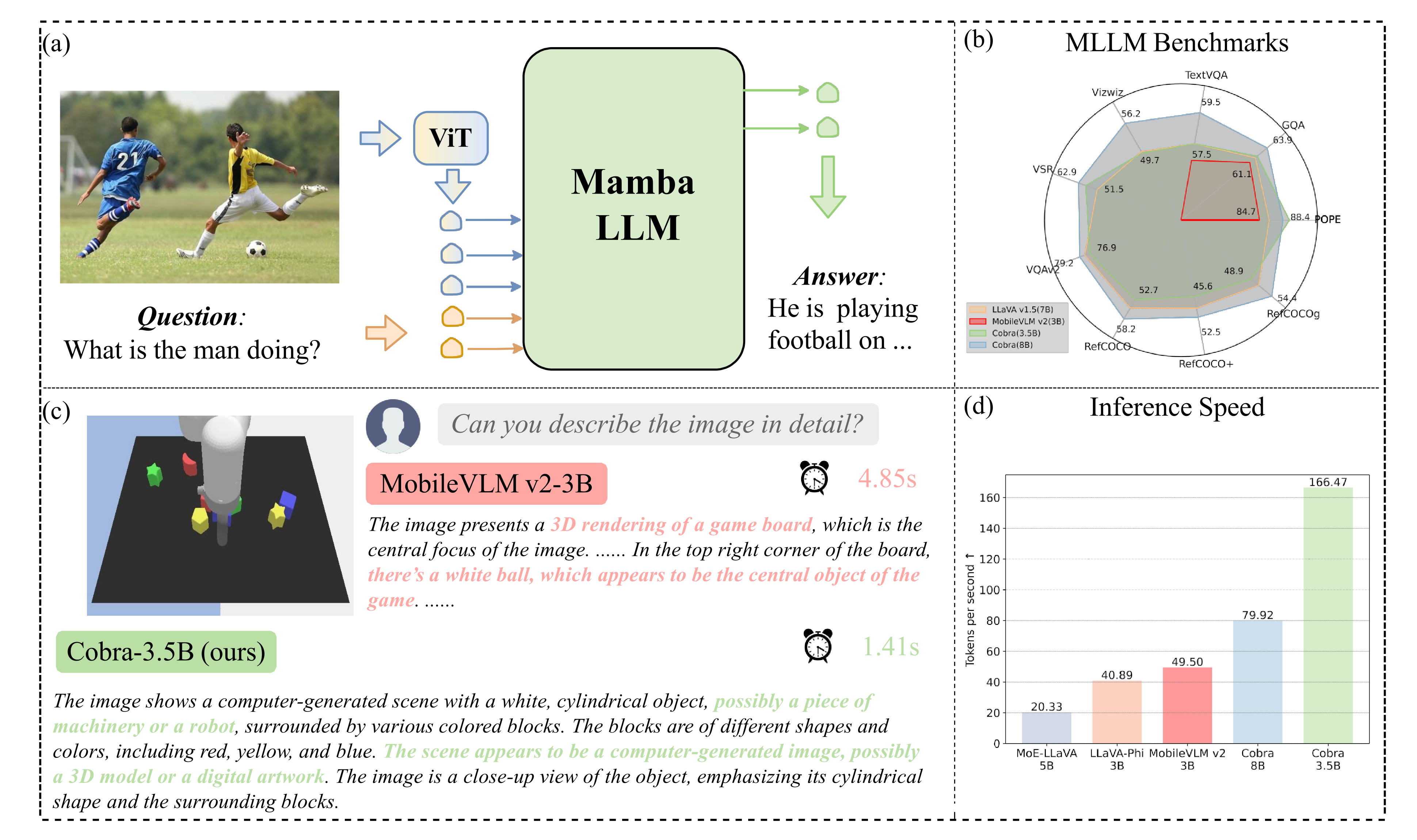}
  \vspace{-2mm}
  \caption{\textbf{Overview of Cobra.} (a) Our innovative integration of a vision encoder with the efficient Mamba language model significantly enhances the reasoning efficiency of MLLMs. (b) Cobra demonstrates competitive performance on general MLLM benchmark tests. (c) Cobra generates accurate textual descriptions (\textit{e.g.}, \green{green} text indicates a correct answer), outperforming current state-of-the-art MLLMs that produce inaccurate answers (shown in \red{red} text) while maintaining rapid reasoning speeds. (d) A comparison of our proposed Cobra and the baseline in terms of the number of tokens generated per second.}
  \label{fig:1}
  \vspace{-2mm}
\end{figure*} 

MLLMs typically rely on the well-known Transformer network as a foundational model for many downstream tasks. However, due to their quadratic computational complexity, Transformer networks are often less efficient, making it challenging to meet the demands of application scenarios that require high real-time performance and are suitable for edge deployment. 
As shown in Figure~\ref{fig:1} (d), the MoE-LLaVA~\cite{lin2024moellava} model can process only 20.33 tokens generated per second, indicating a low processing efficiency. Despite these challenges, there remains a significant demand for MLLMs in such areas. Therefore, \textbf{the ability to deploy MLLMs that support fast inference with low resource utilization is particularly crucial}. 

Traditional approaches have primarily focused on improving the efficiency of MLLMs by reducing model capacity or compressing the length of the visual context while generally maintaining the Transformer architecture within the language model~\cite{zhu2024llavaphi,zhou2024tinyllava,zhang2024tinyllama,chu2023mobilevlm,chu2024mobilevlmv2,lin2024moellava}. For example, LLaVA-Phi~\cite{zhu2024llavaphi} builds a multi-modal base model using a small-scale Phi-2 as the core language model. MobileVLM~\cite{chu2023mobilevlm, chu2024mobilevlmv2} utilizes MobileLLaMA as its base model, training a series of smaller-scale language models based on the LLaMA architecture. These methods aim to enhance the inference speed of MLLMs by reducing the size of the language models. Although this approach improves efficiency, it often comes at the expense of significantly reduced model performance. In Figure~\ref{fig:1} (b), MobileVLM v2-3B~\cite{chu2024mobilevlmv2} has the worst performance on all MLLM benchmarks compared to the other models.

In this paper, our primary goal is to enhance the inference speed of multi-modal large language models (MLLMs) while ensuring their performance remains uncompromising. To achieve this, we propose Cobra, which integrates the Mamba large language model and utilizes the linear scalability of state-space modeling (SSM). This approach effectively addresses the quadratic computational complexity inherent in traditional Transformer architectures. Specifically, Cobra consists of three key components: a vision
encoder that concatenates DINOv2~\cite{oquab2024dinov2} and SigLIP~\cite{zhai2023sigmoid} features, a projector that maps visual features to the language embedding space and the Mamba LLM backbone, as shown in Figure~\ref{fig:1} (a) and Figure~\ref{fig:2}. In Figure~\ref{fig:1} (c), Cobra performs $3 \times \sim 4 \times$ faster than MobileVLM v2-3B. Interestingly, Cobra can generate more accurate responses, as highlighted by the text in green, effectively mitigating the hallucination problem commonly seen in MLLMs. Even compared to the much larger LLaVA v1.5 model with 7 billion parameters, Cobra still performs comparably on several specific benchmarks with about 48\% of the parameters. Our main contributions are summarised:
\begin{itemize}
    \item We investigate that existing MLLMs typically rely on Transformer networks, exhibiting a quadratic computational complexity. To address this inefficiency, we present Cobra, a novel MLLM with linear computation. 
    \item Our research investigates various multi-modal fusion strategies to enhance the integration of visual and language within the Mamba LLM. Extensive experiments evaluate the effectiveness of different fusion approaches.
    \item Extensive experiments are conducted to evaluate the performance of Cobra in comparison to concurrent studies aimed at improving the computational efficiency of foundational MLLMs. Notably, Cobra-3.5B even achieves comparable performance to LLaVA with fewer parameters, underscoring its efficiency. Cobra-8B surpasses the LLaVA v1.5 model of similar size on all tested benchmarks, achieving an average accuracy improvement of approximately 6\%. It also remains faster in inference compared to MobileVLM v2-3B.
\end{itemize}

\section{Related works}

\subsection{Multi-modal Large Language Models}
Building on the success of large language models (LLMs), numerous extensions have been developed to apply LLMs to multi-modal tasks, integrating information from multiple sources such as text, images, and audio to enable comprehensive understanding and reasoning across different modalities~\cite{chu2023mobilevlm,liu2023llava1.5,taori2023alpaca,bai2023qwenvl,alayrac2022flamingo,awadalla2023openflamingo,liu2023llava,chen2023sharegpt4v}. These models leverage vast amounts of data and intricate architectures to achieve state-of-the-art performance in tasks such as image captioning~\cite{ke2019reflective}, visual question answering~\cite{antol2015vqa} and cross-modal retrieval~\cite{hendriksen2023scene}. Recent advances have harnessed the formidable reasoning power of LLMs such as LLaMA~\cite{touvron2023llama} and Vicuna~\cite{vicuna2023}. However, a notable commonality among existing MLLMs is their reliance on the Transformer backbone to model dependencies among sequential tokens. Despite the Transformer network's exceptional capability in capturing relationships within data, its quadratic computational complexity presents a significant drawback, particularly when dealing with large-scale language models. 

To mitigate this problem, several studies have been proposed to present more compact and efficient MLLMs~\cite{zhu2024llavaphi,zhou2024tinyllava,zhang2024tinyllama,chu2023mobilevlm,chu2024mobilevlmv2}. For example, LLaVA-Phi~\cite{zhu2024llavaphi} builds a multi-modal foundation model taking the small-scale Phi-2 as the LLM. MobileVLM~\cite{chu2023mobilevlm,chu2024mobilevlmv2} introduces MobileLLaMA as the base model which trains a family of small-scaled LLM based on LLaMA architecture. However, these methods achieve effective MLLMs by using smaller-scale LLMs that significantly reduce the performance while increasing the speed of inference.

\subsection{State Space Models}

State space models (SSMs) have demonstrated highly promising performance across various tasks, including long-range sequence modeling~\cite{smith2023s5, hasani2022liquid}, image generation~\cite{yan2023diffusion,bellagente2024stable} and reinforcement learning~\cite{bardavid2023decision, lu2023structured}. One of the key advantages of SSMs is their flexibility, as they can be formulated as recurrent neural networks (RNNs) for efficient inference or as models capable of parallel processing entire input sequences, enabling more efficient training. 

Recently, a new selective SSM structure called Mamba~\cite{gu2023mamba} has been introduced, which is regarded as a strong competitor to the Transformer architecture. Compared to LLMs of similar capacity, Mamba-based language models~\cite{dao2024transformersssms,waleffe2024empirical} demonstrate competitive performance with a distinct advantage: their inference speeds scale linearly with sequence length while maintaining constant memory usage. This efficiency allows Mamba to handle long contexts and perform inference more effectively. In contrast, Transformer-based models face challenges such as increasing GPU memory consumption and computation time that grows quadratically with sequence length~\cite{katharopoulos2020transformersrnns}. In this paper, we conduct an in-depth exploration of extending Mamba-based LLMs into practical and efficient MLLMs. Through extensive experiments, we examine the distinctive characteristics of Mamba MLLMs and develop efficient training strategies to significantly enhance their performance.

\section{Methodology}
This section introduces the preliminary concepts of state space models (Section~\ref{sub:pre}). Then we describe the details of our proposed Cobra (Section~\ref{sub:cobra}), which mainly includes the vision encoder, the projector, and the Mamba LLM.

\subsection{Preliminaries}
\label{sub:pre}

Traditional state space models (SSMs)~\cite{gu2022s4, smith2023s5} are characterized by the parameters $(\Delta, A, B, C)$. Given a continuous-time scalar input signal $x(t)$, the SSM can be described by the following ordinary differential equation:
\begin{subequations}
  \begin{align}
        h'(t) & =A h(t)+B x(t), \label{eq:0a} \\
        y(t) & =C h(t), \label{eq:0b}
  \end{align}
\end{subequations}
where the parameters of the SSM, $A \in \mathbb{R}^{N \times N}$, $B \in \mathbb{R}^{N \times 1}$ and $C \in \mathbb{R}^{1 \times N}$, represent constant matrices. The variables $h$, $x$, and $y$ are continuous-time variables concerning time $t$, representing the hidden state, input, and output.

In practice, the SSMs operate in a discretized form to handle input sequences, and we use a mixer layer that constructs an SSM for each input channel independently. $\Delta$ is a time-scale parameter that helps transform $A$ and $B$ into discrete-time parameters $\overline{A}$ and $\overline{B}$, respectively. The discretization rule for $A$ and $B$ with the zero-order hold is as follows:
\begin{subequations}
  \begin{align}
    \overline{A} & = \exp (\Delta A), \\
    \overline{B} & = (\Delta A)^{-1}(\exp (\Delta A)-I) \cdot \Delta B.
  \end{align}
\end{subequations}

Thus, the structured SSM can be summarized as the following recurrence form in Equation\eqref{eq:1a} and \eqref{eq:1b}:
\begin{subequations}
  \begin{align}
    h_k & =\overline{A} h_{k-1}+\overline{B} x_k, \label{eq:1a}\\
    y_k & =C h_k. \label{eq:1b}
  \end{align}
  \label{eq:1ab}
\end{subequations}

The model can also be written as the convolution form \eqref{eq:1c}, \eqref{eq:1d} to process the sequence in parallel:
\begin{subequations}
  \begin{align}
    \overline{{K}} & =\left({C} \overline{{B}}, {C} \overline{{A B}}, \ldots, C \overline{{A}}^k \overline{{B}}, \ldots\right) \label{eq:1c}\\
    y & =x * \overline{{K}} \label{eq:1d}
  \end{align}
  \label{eq:1cd}
\end{subequations}

Based on the structured SSM, the selective SSM~\cite{gu2023mamba} is further introduced to endow the model with the ability to selectively propagate or forget information according to the sequential input tokens. Specifically, the selective SSM achieves these functions by making the parameters ($\overline{A}, \overline{B}, C$) depend on the input $x$, which significantly enhances the model's expressive capacity. Gu et al.~\cite{gu2023mamba} proposed a hardware-aware algorithm called selective scan to allow efficient implementation of the model.

\subsection{Cobra Model}
\label{sub:cobra}

\begin{figure*}[t]
  \centering
  \includegraphics[width=0.9\linewidth]{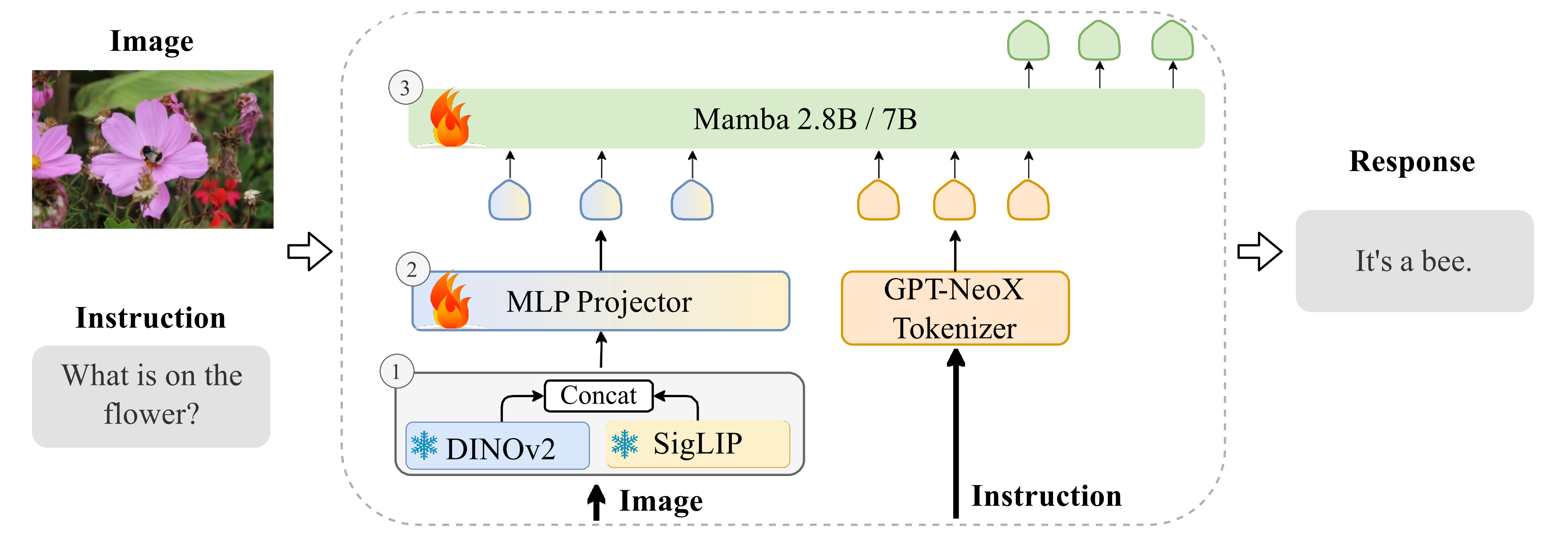}
  \vspace{-2mm}
  \caption{\textbf{Cobra model architecture.} Given an image observation and a language instruction, the model generates the corresponding answer. The architecture consists of three key components: \textcircled{\small{1}} a vision encoder that concatenates DINOv2~\cite{oquab2024dinov2} and SigLIP~\cite{zhai2023sigmoid} features, \textcircled{\small{2}} a projector that maps visual features to the language embedding space and \textcircled{\small{3}} the Mamba LLM backbone, a Mamba 2.8 or 7B-parameter large language model~\cite{gu2023mamba, mercat2024linearizing}.}
  \label{fig:2}
  \vspace{-1mm}
\end{figure*}

To accomplish the purpose of building a multi-modal large language model (MLLM) that is capable of receiving visual information, we introduce Cobra as illustrated in Figure~\ref{fig:2}. Cobra consists of three key components: a vision encoder, a projector, and a Mamba LLM backbone. We present the implementation details for each component below.

\begin{itemize}
 \item Vision encoder: We fuse DINOv2~\cite{oquab2024dinov2} and SigLIP~\cite{zhai2023sigmoid} as our vision backbone. The intuition is that combining the visual representations, which capture low-level spatial properties from DINOv2 and the semantic properties provided by SigLIP further improves the performance on downstream tasks~\cite{tong2024eyes, karamcheti2024prismatic}. Considering an image $X_v \in \mathbb{R}^{C \times H \times W}$ as input, the vision encoder splits the image into $N_v = HW/P^{2}$ same-size patches, where $P^2$ is the patch size. Both two vision encoders take the patchified image as an input token sequence and extract the channel-wise concatenation of the output of two encoders as the compact visual representations $R_v \in \mathbb{R}^{N_v \times (D_{\rm DINOv2} + D_{\rm SigLIP})}$:
  \begin{equation}
      R_v = [\varphi_{\rm DINOv2}(X_v); \varphi_{\rm SigLIP}(X_v)],
  \end{equation}
  for a subsequent task-specific head, where $D_v$ denotes the dimension of the above-produced tokens. 

  \item Projector: The projector is a simple learnable module that aligns the feature of vision and text by transforming the dimension of the original visual representation to the dimension of the tokens in the Mamba language model:
  \begin{equation}
      H_v = \phi(R_v).
  \end{equation}
  We introduce two implementations of the different projectors in Cobra to map visual tokens into the same latent space as the language tokens. The multiple-layer perceptron (MLP) can be used to merge information from different modalities. In addition, the lightweight down-sample projector suggested by~\cite{chu2024mobilevlmv2} is also tested to achieve a greater reduction in computation cost.
  
  \item Mamba backbone: The Mamba backbone is a stack of multiple identical basic blocks with the short convolution, SSM module, the residual connection, and RMSNorm~\cite{zhang2019root} for each block. The model receives the concatenation of visual embeddings transformed from the projection layer and text embeddings, denoted as $H \in \mathbb{R}^{L_{in} \times D}$, and transforms this sequence into target token sequence $Y = \{y_i\}_{i=1}^{L}$ in an auto-regressive manner:
  \begin{equation}
      p(Y | H_v, H_q) = \prod_{i=1}^{L}p(y_i | H_v, H_q, y_{<i}).
  \end{equation}
  Lastly, the tokens will be detokenized to the response answer in natural language.
\end{itemize}

\subsection{Training Recipe}
Recent research~\cite{karamcheti2024prismatic} suggests that the pre-alignment phase may be unnecessary in the LLaVA-based training paradigm~\cite{liu2023llava1.5, chu2024mobilevlmv2} (\textit{i.e.}, training only the pre-alignment phase of the projection layer and fine-tuning the large language model (LLM) for one epoch each). It has been observed that even after fine-tuning, the model remains underfitted. In light of this, we chose to eliminate the pre-alignment stage and instead directly fine-tune the entire LLM backbone and the projector. This fine-tuning process spans two epochs, with random sampling conducted on a combined dataset comprising:

1. The mixed dataset used in LLaVA v1.5, which contains a total of 655K visual multi-turn conversations including academic VQA~\cite{goyal2017making, hudson2019gqa, krishna2016visual, singh2019vqa} samples, as well as visual instruction tuning data in LLaVA-Instruct~\cite{liu2023llava} and pure text instruction tuning data in ShareGPT~\cite{sharegpt}.

2. LVIS-Instruct-4V~\cite{wang2023lvis}, which contains 220K images with visually aligned and context-aware instructions generated by GPT-4V.

3. LRV-Instruct~\cite{liu2023lrv}, a 400K visual instruction dataset that covers 16 vision-and-language tasks aiming on mitigating hallucination.

Overall, the entire dataset contains approximately 1.2 million images, corresponding multi-turn dialogue data, and pure text dialogue data.

\begin{table}[tb]
  \centering
  \begin{tabular}{@{}l@{\hspace{2em}}l@{}}
    \toprule
    Vision Encoder & DINOv2 + SigLIP ViT-SO\\
    LLM init. & Mamba-2.8b-Zephyr / Mamba-7B\\
    Projector init. & Random\\
    Image resolution & $384 \times 384$\\
    Image token num. & 729\\
    Global batch size & 128\\
    Training steps & 19K\\
    Optimizer & AdamW\\
    LR schedule & Cosine decay\\
    Learning Rate & 2e-5\\
    Weight decay & 0.1\\
    Warm-up ratio & 0.03\\
    Number of epochs & 2\\
  \bottomrule
  \end{tabular}
  \vspace{-2mm}
  \caption{The configuration of models and hyperparameters.
  }
  \label{tab:1}
  \vspace{-6mm}
\end{table}

\begin{table*}[tb]
  \centering
  \scalebox{0.93}{
  \begin{tabular}{llccccccc}
    \toprule
    Model & LLM Backbone & \textit{Res.} & \blue{VQA-v2} & \blue{GQA} & \blue{VizWiz} & \blue{TextVQA} & \red{VSR} & \red{POPE}\\
    \midrule
    \multicolumn{9}{c}{\it{\textbf{Large Scale MLLMs}}}   \\
    \midrule
    {OpenFlamingo~\cite{awadalla2023openflamingo}} & {MPT-7B} & {336} & {52.7} & {-} & {27.5} & {33.6} & {-} & {-} \\
    {BLIP-2~\cite{li2023blip2}} & {Vicuna-13B} & {224} & {-} & {41.0} & {19.6} & {42.5} & {50.9} & {-} \\
    {MiniGPT-4~\cite{zhu2023minigpt4}} & {Vicuna-7B} & {224} & {32.2} & {-} & {-} & {-} & {-} & {-} \\
    {InstructBLIP~\cite{dai2023instructblip}} & {Vicuna-7B} & {224} & {-} & {49.2} & {34.5} & {50.1} & \underline{54.3} & {-} \\
    {InstructBLIP~\cite{dai2023instructblip}} & {Vicuna-13B} & {224} & {-} & {49.5} & {33.4} & {50.7} & {52.1} & {-} \\
    {Shikra~\cite{chen2023shikra}} & {Vicuna-13B} & {224} & {77.4} & {-} & {-} & {-} & {-} & {-} \\
    {IDEFICS~\cite{laurencon2023obelics}} & {LLaMA-7B} & {224} & {50.9} & {-} & {35.5} & {25.9} & {-} & {-} \\
    {IDEFICS~\cite{laurencon2023obelics}} & {LLaMA-75B} & {224} & {60.0} & {-} & {36.0} & {30.9} & {-} & {-} \\
    {Qwen-VL~\cite{bai2023qwenvl}} & {Qwen-7B} & {448} & {78.2} & {59.3} & {35.2} & {63.8} & {-} & {-} \\
    {LLaVA v1.5~\cite{liu2023llava1.5}} & {Vicuna-7B} & {336} & \underline{78.5} & \underline{62.0} & \underline{50.0} & \underline{58.2} & {51.5} & \underline{85.9} \\
    \midrule
    \textbf{Cobra-8B (ours)} & Mamba-7B & {384} & \textbf{79.2} & \textbf{63.9} & \textbf{56.2} & \textbf{59.5} & \textbf{62.9} & \textbf{87.6} \\
    \midrule
    \multicolumn{9}{c}{\it{\textbf{Small Scale MLLMs}}}  \\
    \midrule
    {MoE-LLaVA~\cite{lin2024moellava}} & {StableLM-1.6B} & {336} & {76.7} & {60.3} & {36.2} & {50.1} & {-} & \underline{85.7} \\
    {MoE-LLaVA~\cite{lin2024moellava}} & {Phi2-2.7B} & {384} & \textbf{79.9} & \textbf{62.6} & \underline{43.7} & {57.0} & {-} & \underline{85.7} \\
    LLaVA-Phi~\cite{zhu2024llavaphi} & Phi2-2.7B & {336} & 71.4 & - & 35.9 & 48.6 & - & 85.0\\
    MobileVLM v2~\cite{chu2024mobilevlmv2} & MobileLLaMA-2.7B & {336} & - & 61.1 & - & \underline{57.5} & - & 84.7\\
    \midrule
    \textbf{Cobra-3.5B (ours)} & Mamba-2.8B & {384} & \underline{77.8} & \underline{62.3} & \textbf{49.7} & \textbf{58.2} & \textbf{58.4} & \textbf{88.4}\\
    \bottomrule
  \end{tabular}}
  \vspace{-2.5mm}
  \caption{Experiments of four open-ended benchmarks (\blue{blue}) and two closed-set benchmarks (\red{red}). \textit{Res.} represents the image resolution used for the vision encoder input. The best performance is highlighted in bold and the second-best result is underlined.}
  \label{tab:2}
  \vspace{-2mm}
\end{table*}

\section{Experiments}
In this section, we conduct extensive experiments to evaluate the performance of our proposed Cobra method, aiming to answer the following questions: 
\textbf{RQ1:} How does the performance of our proposed Cobra method compare with state-of-the-art MLLMs? (Section~\ref{subsec:per}) \textbf{RQ2:} How does the inference speed of Cobra compare to three transformer-based baselines? (Section~\ref{subsec:inf})
\textbf{RQ3:} How effective is the proposed Cobra in different settings (or ablation study)? (Section~\ref{subsec:abla})

\subsection{Experimental Setup}

\noindent \textbf{Datasets.}
We conduct our experiments on a diverse set of nine benchmarks, including (1) four open-ended visual question answering (VQA), \textit{i.e.}, VQA-v2~\cite{goyal2017making}, GQA~\cite{hudson2019gqa}, VizWiz~\cite{gurari2018vizwiz} and TextVQA~\cite{singh2019vqa}. (2) two closed-set visual question answering (VQA), \textit{i.e.}, VSR~\cite{liu2023visual} and POPE~\cite{li2023evaluating}. (3) three visual grounding, \textit{i.e.}, RefCOCO, RefCOCO+ and RefCOCOg~\cite{Kazemzadeh2014ReferItGame, yu2016modeling}. 

VQA-v2~\cite{goyal2017making} evaluates models’ general ability to understand and reason about images and questions. GQA~\cite{hudson2019gqa} assesses spatial understanding and multi-step inference in real-world images. VizWiz~\cite{gurari2018vizwiz} is similar to VQA-v2 but includes a series of unanswerable questions. 
TextVQA~\cite{singh2019vqa} focuses on reasoning from text in images.
VSR~\cite{liu2023visual} is composed of demanding True/False questions that probe individual spatial relationships within various scenes, which is challenging to MLLMs.
POPE~\cite{li2023evaluating} is comprised of specific Yes/No questions designed to evaluate MLLMs' tendency to generate hallucinations. RefCOCO focuses on short descriptions with spatial anchors, RefCOCO+ relies on appearance-based descriptions, and RefCOCOg emphasizes long and rich descriptions~\cite{Kazemzadeh2014ReferItGame, yu2016modeling}.

\noindent \textbf{Baseline methods.}
We compare Cobra to a large number of algorithms that span different sizes, including (1) large-scale MLLMs: OpenFlamingo~\cite{awadalla2023openflamingo}, BLIP-2~\cite{li2023blip2}, MiniGPT-4~\cite{zhu2023minigpt4}, InstructBLIP~\cite{dai2023instructblip}, Shikra~\cite{chen2023shikra}, IDEFICS~\cite{laurencon2023obelics}, Qwen-VL~\cite{bai2023qwenvl} and LLaVA v1.5~\cite{liu2023llava1.5}.
(2) small-scale MLLMs: MoE-LLaVA~\cite{lin2024moellava}, LLaVA-Phi~\cite{zhu2024llavaphi} and MobileVLM v2~\cite{chu2024mobilevlmv2}.

\noindent \textbf{Implementation details.}
Our training process includes multi-modal instruction tuning, during which we fine-tune both the multi-modal projector and the Mamba LLM. The model is trained using 8 NVIDIA A100 80GB GPUs. We have selected various open-source model weights, including Mamba with 2.8 billion and 7 billion parameters, to serve as the LLM backbone for our model. The model configurations and hyperparameters are detailed in Table~\ref{tab:1}, with additional information provided in the supplementary material.

\begin{table*}[t]
\setlength{\tabcolsep}{1mm}
  \centering
  {
  \begin{tabular}{l*{6}{l}}
    \toprule
    Model & LLM Backbone & Total Params & Visual Tokens & \makecell[c]{$Eval_{avg}$ (tokens/s)} & $Total$ (s) \\
    \midrule
    \multirow{1}{*}{\makecell[c]{MoE-LLaVA}}
            & Phi-2-2.7B & {5.3B (3.6B Activated)} & {576} & {20.33} & {12.59}  \\
    \multirow{1}{*}{\makecell[c]{LLaVA-Phi}}
            & Phi-2-2.7B & {3.1B} & {576} & {40.89} & {6.26}  \\
    \multirow{1}{*}{\makecell[c]{MobileVLM v2}}
            & MobileLLaMA-2.7B & {3.1B} & {144} & 49.50 & 5.17  \\
    \midrule
    \multirow{1}{*}{\makecell[c]{\textbf{Cobra-3.5B (ours)}}}
            & Mamba-2.8B & {3.5B} & {729} & \textbf{166.47} & \textbf{1.54} \\
    \multirow{1}{*}{\makecell[c]{\textbf{Cobra-LDPv2-3.5B (ours)}}}
            & Mamba-2.8B & {3.5B} & {196} & \textbf{166.85} & \textbf{1.53} \\
    \multirow{1}{*}{\makecell[c]{\textbf{Cobra-8B (ours)}}}
            & Mamba-7B & {7.8B} & {729} & \textbf{79.92} & \textbf{3.20} \\
    \bottomrule
  \end{tabular}}
  \vspace{-2mm}
  \caption{Latency comparison of small-scale MLLMs with $\sim$3 billion parameters.}
  \label{tab:4}
  \vspace{-4mm}
\end{table*}

\noindent \textbf{Prompt order.}
In our prompt template design, we were surprised to discover that the word order in the templates significantly impacts the model's performance, particularly in TextVQA. For example, Cobra, which follows LLaVA and InstructBLIP evaluations, uses input tokens parsed by the OCR system as prompts—formatted as \textbf{``Question{$\backslash$n} Reference OCR token: ...”}. We found that this specific prompt structure reduced performance substantially, from 47.9\% to 43.0\%, compared to not using any prompts at all. Through extensive experimental exploration, we addressed this issue by adjusting the prompt order to \textbf{``Reference OCR token: ...{$\backslash$n} Question”}, which improved performance. This sensitivity to prompt order may be due to an inductive bias in the RNN model. We hope that our findings will encourage further research in the community on this problem.

\begin{table}[tb]
\setlength{\tabcolsep}{1mm}
  \centering
  {
  \begin{tabular}{lcccc}
    \toprule
    Model & {RefCOCO} & {RefCOCO+} & {RefCOCOg} & \textit{Avg.}\\
    \midrule
    {LLaVA v1.5} & {55.1} & {49.5} & {50.9} & {51.8}\\
    \midrule
    \textbf{Cobra-3.5B} & {52.7} & {45.6} & {46.9} & {48.4}\\
    \textbf{Cobra-8B} & \textbf{58.2} & \textbf{52.5} & \textbf{54.4} & \textbf{55.0}\\
    \bottomrule
  \end{tabular}}
  \vspace{-1mm}
  \caption{Experiments of three visual grounding benchmarks. \textit{Avg.} represents the average accuracy of the model on three benchmarks. The best performance is highlighted in bold.}
  \label{tab:3}
  \vspace{-4mm}
\end{table}

\subsection{Overall Performance}
\label{subsec:per}

In Table~\ref{tab:1}, we report the overall performance of Cobra and fourteen baselines under large-scale MLLMs and small-scale MLLMs on six datasets. According to Table~\ref{tab:1}, we have the following findings: (1) \textbf{In the large scale MLLMs setting with more than 7 billion parameters}. our proposed Cobra-8B achieves the best performance on all evaluated benchmarks. (2) \textbf{In the small scale MLLMs setting with around 3 billion (total or activated) parameters}. Cobra-3.5B achieved the best performance on all benchmarks except VQA-v2 and GQA, where it was only surpassed by MoELLAVA, a multi-modal mixture-of-experts language model expanded and fine-tuned from phi-2-2.7B. Our model lags behind by 1\%-2\% in accuracy on these two metrics, while our inference speed is over 8 times faster than that of the model as shown in Section~\ref{subsec:inf}. 

It is noteworthy that Cobra-3.5B, with only 48\% of the total parameters of LLaVA v1.5-7B, achieves comparable results on open-ended VQA benchmarks and shows significant improvements in the challenging closed-set prediction tasks of VSR and POPE. On these two benchmarks, there are performance improvements of 6.9\% and 2.5\% respectively.

As shown in Table~\ref{tab:3}, we also evaluated the localization capabilities of our two models alongside LLaVA v1.5-7B. The results indicate that Cobra-3.5B has accuracy rates that are 3\%-4\% lower than LLaVA v1.5-7B across all three benchmarks. In contrast, Cobra-8B exhibits the highest accuracy among the three models, outperforming the others by over 3\% in accuracy on all benchmarks. Given that the training schemes for Cobra were identical, these results demonstrate that the grounding ability of the model is significantly influenced by the performance of the language model itself.

\begin{table*}[t]
  \centering
  \scalebox{0.85}{
  \begin{tabular}{l*{10}{c}}
    \toprule
    Model & VQA-v2 & GQA & VizWiz & TextVQA & VSR & POPE & RefCOCO & RefCOCO+ & RefCOCOg \\
    \midrule
            & {77.8} & {62.3} & {49.7} & {58.2} & {58.4} & {88.4} & {52.7} & {45.6} & {48.9}  \\
    \multirow{1}{*}{\makecell[c]{w/ SigLIP}}
            & {77.5 (\red{0.3 $\downarrow$})} & {61.8 (\red{0.5 $\downarrow$})} & {48.3 (\red{1.4 $\downarrow$})} & {58.8 (\green{0.6 $\uparrow$})} & {53.2 (\red{5.2 $\downarrow$})} & {88.2 (\red{0.2 $\downarrow$})} & {46.7 (\red{6.0 $\downarrow$})} & {40.1 (\red{5.5 $\downarrow$})} & {43.8 (\red{5.1 $\downarrow$})}  \\
    \multirow{1}{*}{\makecell[c]{w/ LDPv2}}
            & {76.2 (\red{1.6 $\downarrow$})} & {61.9 (\red{0.4 $\downarrow$})} & {50.2 (\green{0.5 $\uparrow$})} & {54.7 (\red{3.5 $\downarrow$})} & {56.1 (\red{2.3 $\downarrow$})} & {87.7 (\red{0.7 $\downarrow$})} & {50.3 (\red{2.4 $\downarrow$})} & {42.9 (\red{2.7 $\downarrow$})} & {46.9 (\red{2.0 $\downarrow$})}  \\
    \multirow{1}{*}{\makecell[c]{w/ Base}}
            & {77.8 (0.0 $\updownarrow$)} & {62.7 (\green{0.4 $\uparrow$})} & {47.2 (\red{2.5 $\downarrow$})} & {57.9 (\red{0.3 $\downarrow$})} & {54.4 (\red{4.0 $\downarrow$})} & {89.0 (\green{0.6 $\uparrow$})} & {52.2 (\red{0.5 $\downarrow$})} & {45.6 (0.0 $\updownarrow$)} & {48.6 (\red{0.3 $\downarrow$})}  \\
    \multirow{1}{*}{\makecell[c]{w/ 1 Ep FT}}
            & {76.5 (\red{1.3 $\downarrow$})} & {60.9 (\red{1.4 $\downarrow$})} & {48.5 (\red{1.2 $\downarrow$})} & {57.5 (\red{0.7 $\downarrow$})} & {53.8 (\red{4.6 $\downarrow$})} & {88.1 (\red{0.3 $\downarrow$})} & {42.5 (\red{10.2 $\downarrow$})} & {34.3 (\red{11.3 $\downarrow$})} & {39.0 (\red{9.9 $\downarrow$})}  \\
    \multirow{1}{*}{\makecell[c]{w/ PT+FT}}
            & {75.7 (\red{2.1 $\downarrow$})} & {60.4 (\red{1.9 $\downarrow$})} & {44.2 (\red{5.5 $\downarrow$})} & {58.0 (\red{0.2 $\downarrow$})} & {51.6 (\red{6.8 $\downarrow$})} & {86.9 (\red{1.5 $\downarrow$})} & {37.3 (\red{15.4 $\downarrow$})} & {29.7 (\red{15.9 $\downarrow$})} & {34.3 (\red{14.6 $\downarrow$})}  \\
    \bottomrule
  \end{tabular}}
  \vspace{-2mm}
  \caption{Ablation studies of Cobra-3.5B on vision encoders, projectors, language models and training strategies}
  \label{tab:5}
  \vspace{-2mm}
\end{table*}

\begin{table*}[t]
    \centering 
    \scalebox{0.87}{
      \begin{tabular}{p{4cm}  p{7.5cm}  p {7.5cm}}
        \toprule
        {\bf Visual input example} & {\bf Spatial Reasoning} & {\bf Scene Description} \\
        \midrule
        &  \includegraphics[height=3cm]{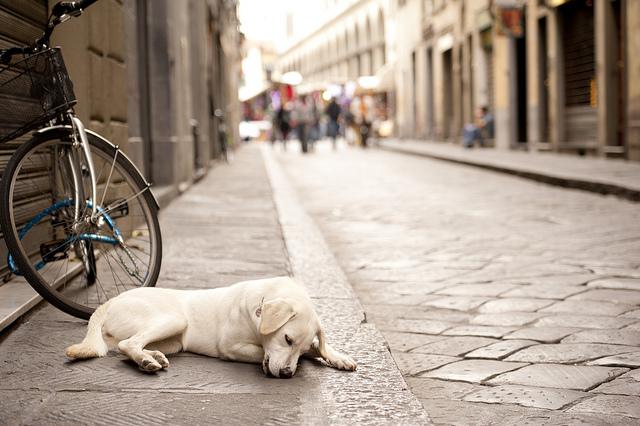}
        &  \includegraphics[height=3cm]{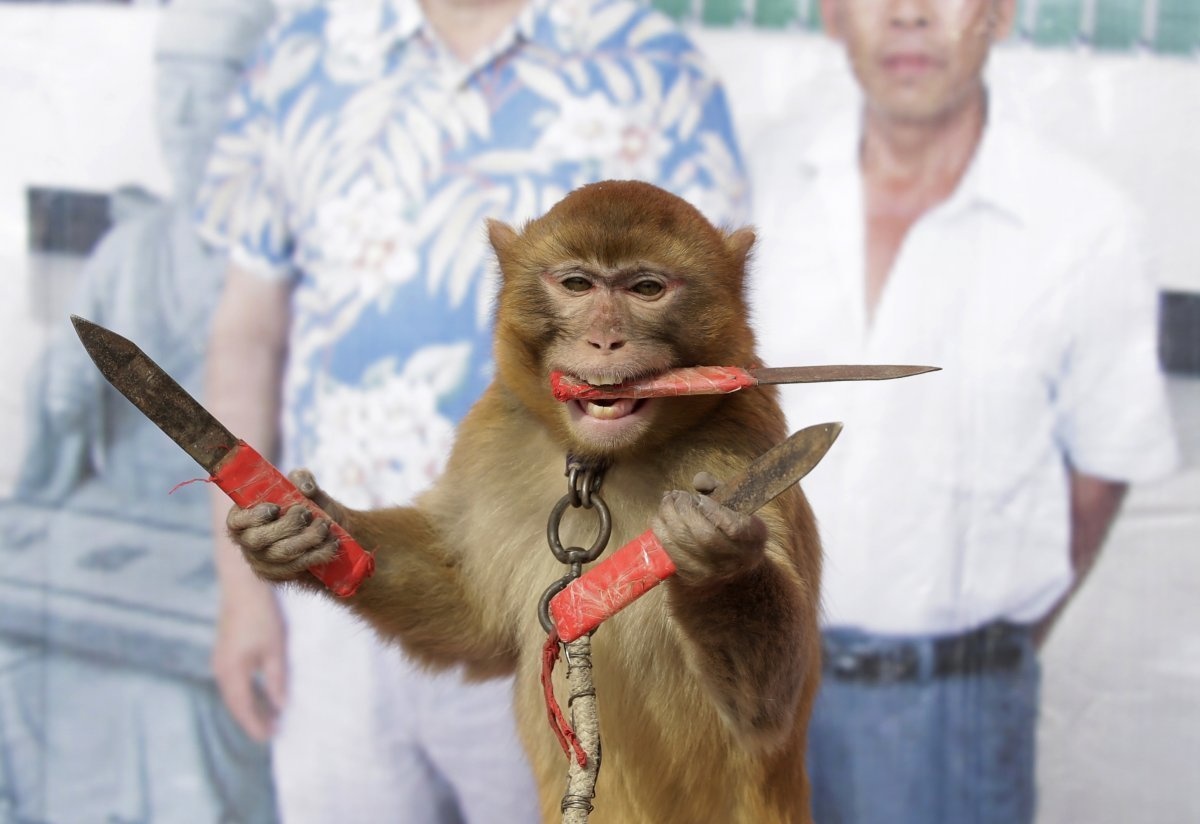}
        \\
        User & Is the bicycle parked to the right of the dog? & What's going on in this image? \\
        \midrule
        Cobra-3.5B (ours) & The bicycle is actually parked on the \green{left} side of the dog, \green{not the right}. & A monkey is holding two knives in its hands, \textcolor{blue}{while a man in the background is wearing a white shirt with a floral pattern}.\\
        \midrule
        LLaVA v1.5-7B & \red{Yes}, the bicycle is parked on the \red{right} side of the dog. & In this image, a monkey is holding two knives in its hands, seemingly posing for a picture. \\
        \bottomrule
      \end{tabular}}
    \vspace{-2mm}
    \caption{\textbf{Visualization of the MLLM example.} Cobra generates accurate and more detailed textual descriptions compared with the baseline, where \green{green} indicates a correct answer, \red{red} produces inaccurate answers and \textcolor{blue}{blue} is a more detailed description.}
    \label{tab:6}  
    \vspace{-4mm}
\end{table*}

\subsection{Inference Speed}
\label{subsec:inf}

We evaluated the generation speed of our model compared to three transformer-based baseline models of similar activated parameter scales with different architectures: MoE-LLaVA, LLaVA-Phi, and MobileVLM v2.

In the evaluation, all models received the same example image. We used the same question ``\textit{Describe the image specifically}" as the textual prompt and set the number of output tokens to 256 for all models. The total time $T_{total}$ from the image encoding to finished generating the complete answer is recorded and we calculated the average number of tokens generated per second by $Eval_{avg} = 256 / T_{total}$.

All the evaluations were done on the hardware with a single Nvidia A100 PCIe 80GB GPU. The results from Table~\ref{tab:4} show that our model has a significant advantage in inference speed compared to transformer-based models. Compared to MobileVLM v2, which has undergone several lightweight optimizations, Cobra only took about 30\% of the time to complete inference when the number of visual tokens processed significantly increased.

We also evaluated the results of Cobra-LDPv2, a variant of our model that replaces the projector with an LDPv2 block, which reduces the number of visual tokens per image to 196. The results showed no significant speed improvement in our evaluation method. Due to the nature of parallel RNN models, the number of prompt tokens only affects the speed of the model's first parallel forward process. Given that LDP significantly compresses visual information through pooling, it can impact the performance of MLLM to some extent (see our ablation studies for performance comparison). We believe that for the structure of Cobra or other RNN-based MLLMs, adopting such a lightweight design on the projector may be unnecessary.

\subsection{Ablation Studies}
\label{subsec:abla}

We conduct ablation studies to verify the network design of Cobra, mainly involving the choice of projectors, vision encoders, language models, and training strategies. 

\noindent \textbf{Vision encoders.}
Recent works discover that despite CLIP-like language-image models may offer rich semantics, it has the potential to lose the detailed information for images themselves. Therefore, we adopt DINOv2 as a supplementary encoder and concatenate visual representations from two encoders for the subsequent LLM. As shown in Table~\ref{tab:4}, the fusion of DINOv2 and SigLIP features leads to better performance compared with SigLIP-only on all the benchmarks except TextVQA. Especially, we found the fused architecture significantly improves the accuracy by 5\%-6\% on VSR and localization benchmarks. This result implies that there is a meaningful principle when selecting the vision encoder for downstream tasks.

\noindent \textbf{Projectors.}
Besides, a different choice of projection layer is used in the experiments. We investigate a lightweight down-sample projector (LDPv2) to see if we can further speed up the inference process without obvious deterioration in performance. Applying LDPv2 to Cobra harms the performance on all benchmarks except VizWiz. Unfortunately, we observed that the models using LDPv2 show a significant decrease in accuracy on TextVQA, VSR, and localization benchmarks, which require precise visual information.

\noindent \textbf{Base or instruct-tuned LLMs.}
We also explored the application of different Mamba LLMs. Specifically, we chose a base model that had not been fine-tuned on any chat datasets. As indicated in Table~\ref{tab:4}, the fine-tuned model achieved notably higher accuracy on the VizWiz and VSR benchmarks compared to the pre-trained model that did not utilize chat corpora, with accuracy improvements of 2.5\% and 4\%, respectively. On other benchmarks, the differences were not significant with accuracy gaps within 1\%. The chat model exhibits a slight disadvantage compared to the base model only on the GQA and POPE benchmarks.

\noindent \textbf{Training strategies.}
Different training strategies were investigated. The results show that fine-tuning the language model for two epochs yields strictly better performance on all evaluated benchmarks compared with the model that only fine-tuned for one epoch. This suggests that the model may be underfitted with only one epoch of training. 

Additionally, we discovered that initializing a pre-aligned projector during the fine-tuning stage actually harms the model's performance, resulting in consistently lower accuracy across all benchmarks compared to a model with a randomly initialized projector (when both models are fine-tuned for one epoch) except TextVQA. This conclusion differs from several different approaches that treat pre-alignment as the first stage of training.~\cite{lin2024moellava, chu2024mobilevlmv2}.

\noindent \textbf{Visualization of the MLLM example.}
We visualize some examples to demonstrate the performance. In Table~\ref{tab:6}, Cobra outperforms LLaVA v1.5 in the first example involving the judgment of spatial relationships. Cobra correctly identified that the dog was parked to the right of the bicycle, whereas LLaVA provided the opposite, incorrect answer. In the second example, Cobra offered a more detailed description of the background information compared with LLaVA. More examples are shown in the supplementary material.

\section{Conclusion}
In this study, we propose Cobra, which addresses the efficiency bottleneck of existing multi-modal large language models (MLLMs) that rely on Transformer networks with quadratic computational complexity. We explore combining language models with linear computational complexity and multi-modal inputs. In terms of fusing visual and linguistic information, we have successfully optimized the internal information integration of the Mamba language model through in-depth research on different modality fusion schemes, achieving more effective multi-modal representation. Experiments demonstrate that Cobra not only significantly improves computational efficiency, but also performs comparably to advanced models like LLaVA, especially excelling in overcoming visual hallucination and spatial relationship judgment. It even significantly reduces the number of parameters. This opens up new possibilities for deploying high-performance AI models in environments that require high-frequency processing of visual information, such as visual-based robotic feedback control, in the future. 

\section{Acknowledgments}
This work was supported by the National Science and Technology Innovation 2030 - Major Project (Grant No. 2022ZD0208800), and NSFC General Program (Grant No. 62176215).

\bibliography{aaai25}

\begin{thebibliography}{67}
\providecommand{\natexlab}[1]{#1}

\bibitem[{Alayrac et~al.(2022)Alayrac, Donahue, Luc, Miech, Barr et~al.}]{alayrac2022flamingo}
Alayrac, J.-B.; Donahue, J.; Luc, P.; Miech, A.; Barr, I.; et~al. 2022.
\newblock Flamingo: a Visual Language Model for Few-Shot Learning.
\newblock arXiv:2204.14198.

\bibitem[{Antol et~al.(2015)Antol, Agrawal, Lu, Mitchell, Batra, Zitnick, and Parikh}]{antol2015vqa}
Antol, S.; Agrawal, A.; Lu, J.; Mitchell, M.; Batra, D.; Zitnick, C.~L.; and Parikh, D. 2015.
\newblock Vqa: Visual question answering.
\newblock In \emph{Proceedings of the IEEE international conference on computer vision}, 2425--2433.

\bibitem[{Awadalla et~al.(2023)Awadalla, Gao, Gardner, Hessel, Hanafy et~al.}]{awadalla2023openflamingo}
Awadalla, A.; Gao, I.; Gardner, J.; Hessel, J.; Hanafy, Y.; et~al. 2023.
\newblock OpenFlamingo: An Open-Source Framework for Training Large Autoregressive Vision-Language Models.
\newblock arXiv:2308.01390.

\bibitem[{Bai et~al.(2023)Bai, Bai, Yang, Wang, Tan et~al.}]{bai2023qwenvl}
Bai, J.; Bai, S.; Yang, S.; Wang, S.; Tan, S.; et~al. 2023.
\newblock Qwen-VL: A Versatile Vision-Language Model for Understanding, Localization, Text Reading, and Beyond.
\newblock arXiv:2308.12966.

\bibitem[{Bar-David et~al.(2023)Bar-David, Zimerman, Nachmani, and Wolf}]{bardavid2023decision}
Bar-David, S.; Zimerman, I.; Nachmani, E.; and Wolf, L. 2023.
\newblock Decision S4: Efficient Sequence-Based RL via State Spaces Layers.
\newblock arXiv:2306.05167.

\bibitem[{Bellagente et~al.(2024)Bellagente, Tow, Mahan, Phung, Zhuravinskyi, Adithyan, Baicoianu, Brooks, Cooper, Datta, Lee, Mostaque, Pieler, Pinnaparju, Rocha, Saini, Teufel, Zanichelli, and Riquelme}]{bellagente2024stable}
Bellagente, M.; Tow, J.; Mahan, D.; Phung, D.; Zhuravinskyi, M.; Adithyan, R.; Baicoianu, J.; Brooks, B.; Cooper, N.; Datta, A.; Lee, M.; Mostaque, E.; Pieler, M.; Pinnaparju, N.; Rocha, P.; Saini, H.; Teufel, H.; Zanichelli, N.; and Riquelme, C. 2024.
\newblock Stable LM 2 1.6B Technical Report.
\newblock arXiv:2402.17834.

\bibitem[{Brohan et~al.(2023)Brohan, Brown, Carbajal, Chebotar, Chen et~al.}]{brohan2023rt2}
Brohan, A.; Brown, N.; Carbajal, J.; Chebotar, Y.; Chen, X.; et~al. 2023.
\newblock RT-2: Vision-Language-Action Models Transfer Web Knowledge to Robotic Control.
\newblock arXiv:2307.15818.

\bibitem[{Chen et~al.(2023{\natexlab{a}})Chen, Zhang, Zeng, Zhang, Zhu et~al.}]{chen2023shikra}
Chen, K.; Zhang, Z.; Zeng, W.; Zhang, R.; Zhu, F.; et~al. 2023{\natexlab{a}}.
\newblock Shikra: Unleashing Multimodal LLM's Referential Dialogue Magic.
\newblock arXiv:2306.15195.

\bibitem[{Chen et~al.(2023{\natexlab{b}})Chen, Li, Dong, Zhang, He, Wang et~al.}]{chen2023sharegpt4v}
Chen, L.; Li, J.; Dong, X.; Zhang, P.; He, C.; Wang, J.; et~al. 2023{\natexlab{b}}.
\newblock ShareGPT4V: Improving Large Multi-Modal Models with Better Captions.
\newblock arXiv:2311.12793.

\bibitem[{Chiang et~al.(2023)Chiang, Li, Lin, Sheng, Wu, Zhang, Zheng, Zhuang, Zhuang, Gonzalez, Stoica, and Xing}]{vicuna2023}
Chiang, W.-L.; Li, Z.; Lin, Z.; Sheng, Y.; Wu, Z.; Zhang, H.; Zheng, L.; Zhuang, S.; Zhuang, Y.; Gonzalez, J.~E.; Stoica, I.; and Xing, E.~P. 2023.
\newblock Vicuna: An Open-Source Chatbot Impressing GPT-4 with 90\%* ChatGPT Quality.

\bibitem[{Chu et~al.(2023)Chu, Qiao, Lin, Xu, Yang, Hu et~al.}]{chu2023mobilevlm}
Chu, X.; Qiao, L.; Lin, X.; Xu, S.; Yang, Y.; Hu, Y.; et~al. 2023.
\newblock MobileVLM : A Fast, Strong and Open Vision Language Assistant for Mobile Devices.
\newblock arXiv:2312.16886.

\bibitem[{Chu et~al.(2024)Chu, Qiao, Zhang, Xu, Wei, Yang et~al.}]{chu2024mobilevlmv2}
Chu, X.; Qiao, L.; Zhang, X.; Xu, S.; Wei, F.; Yang, Y.; et~al. 2024.
\newblock MobileVLM V2: Faster and Stronger Baseline for Vision Language Model.
\newblock arXiv:2402.03766.

\bibitem[{Cui et~al.(2023)Cui, Yuan, Ding, Yao, Zhu, Ni et~al.}]{cui2023ultrafeedback}
Cui, G.; Yuan, L.; Ding, N.; Yao, G.; Zhu, W.; Ni, Y.; et~al. 2023.
\newblock UltraFeedback: Boosting Language Models with High-quality Feedback.
\newblock arXiv:2310.01377.

\bibitem[{Dai et~al.(2023)Dai, Li, Li, Tiong, Zhao et~al.}]{dai2023instructblip}
Dai, W.; Li, J.; Li, D.; Tiong, A. M.~H.; Zhao, J.; et~al. 2023.
\newblock InstructBLIP: Towards General-purpose Vision-Language Models with Instruction Tuning.
\newblock arXiv:2305.06500.

\bibitem[{Dao and Gu(2024)}]{dao2024transformersssms}
Dao, T.; and Gu, A. 2024.
\newblock Transformers are SSMs: Generalized Models and Efficient Algorithms Through Structured State Space Duality.
\newblock arXiv:2405.21060.

\bibitem[{Ding et~al.(2023)Ding, Chen, Xu, Qin, Zheng, Hu et~al.}]{ding2023enhancing}
Ding, N.; Chen, Y.; Xu, B.; Qin, Y.; Zheng, Z.; Hu, S.; et~al. 2023.
\newblock Enhancing Chat Language Models by Scaling High-quality Instructional Conversations.
\newblock arXiv:2305.14233.

\bibitem[{Ding et~al.(2024)Ding, Zhao, Song, Zhang, Zhang et~al.}]{ding2024quarvla}
Ding, P.; Zhao, H.; Song, W.; Zhang, W.; Zhang, M.; et~al. 2024.
\newblock QUAR-VLA: Vision-Language-Action Model for Quadruped Robots.
\newblock arXiv:2312.14457.

\bibitem[{Gao et~al.(2023)Gao, Han, Zhang, Lin, Geng et~al.}]{gao2023llamaadapter}
Gao, P.; Han, J.; Zhang, R.; Lin, Z.; Geng, S.; et~al. 2023.
\newblock LLaMA-Adapter V2: Parameter-Efficient Visual Instruction Model.
\newblock arXiv:2304.15010.

\bibitem[{Goyal et~al.(2017)Goyal, Khot, Summers-Stay, Batra, and Parikh}]{goyal2017making}
Goyal, Y.; Khot, T.; Summers-Stay, D.; Batra, D.; and Parikh, D. 2017.
\newblock Making the V in VQA Matter: Elevating the Role of Image Understanding in Visual Question Answering.
\newblock arXiv:1612.00837.

\bibitem[{Gu and Dao(2023)}]{gu2023mamba}
Gu, A.; and Dao, T. 2023.
\newblock Mamba: Linear-Time Sequence Modeling with Selective State Spaces.
\newblock arXiv:2312.00752.

\bibitem[{Gu, Goel, and Ré(2022)}]{gu2022s4}
Gu, A.; Goel, K.; and Ré, C. 2022.
\newblock Efficiently Modeling Long Sequences with Structured State Spaces.
\newblock arXiv:2111.00396.

\bibitem[{Gurari et~al.(2018)Gurari, Li, Stangl, Guo, Lin et~al.}]{gurari2018vizwiz}
Gurari, D.; Li, Q.; Stangl, A.~J.; Guo, A.; Lin, C.; et~al. 2018.
\newblock VizWiz Grand Challenge: Answering Visual Questions from Blind People.
\newblock arXiv:1802.08218.

\bibitem[{Hasani et~al.(2022)Hasani, Lechner, Wang, Chahine, Amini, and Rus}]{hasani2022liquid}
Hasani, R.; Lechner, M.; Wang, T.-H.; Chahine, M.; Amini, A.; and Rus, D. 2022.
\newblock Liquid Structural State-Space Models.
\newblock arXiv:2209.12951.

\bibitem[{Hendriksen et~al.(2023)Hendriksen, Vakulenko, Kuiper, and de~Rijke}]{hendriksen2023scene}
Hendriksen, M.; Vakulenko, S.; Kuiper, E.; and de~Rijke, M. 2023.
\newblock Scene-centric vs. object-centric image-text cross-modal retrieval: a reproducibility study.
\newblock In \emph{European Conference on Information Retrieval}, 68--85. Springer.

\bibitem[{Hudson and Manning(2019)}]{hudson2019gqa}
Hudson, D.~A.; and Manning, C.~D. 2019.
\newblock GQA: A New Dataset for Real-World Visual Reasoning and Compositional Question Answering.
\newblock arXiv:1902.09506.

\bibitem[{Karamcheti et~al.(2024)Karamcheti, Nair, Balakrishna, Liang, Kollar et~al.}]{karamcheti2024prismatic}
Karamcheti, S.; Nair, S.; Balakrishna, A.; Liang, P.; Kollar, T.; et~al. 2024.
\newblock Prismatic VLMs: Investigating the Design Space of Visually-Conditioned Language Models.
\newblock arXiv:2402.07865.

\bibitem[{Katharopoulos et~al.(2020)Katharopoulos, Vyas, Pappas, and Fleuret}]{katharopoulos2020transformersrnns}
Katharopoulos, A.; Vyas, A.; Pappas, N.; and Fleuret, F. 2020.
\newblock Transformers are RNNs: Fast Autoregressive Transformers with Linear Attention.
\newblock arXiv:2006.16236.

\bibitem[{Kazemzadeh et~al.(2014)Kazemzadeh, Ordonez, andre Matten, and Berg}]{Kazemzadeh2014ReferItGame}
Kazemzadeh, S.; Ordonez, V.; andre Matten, M.; and Berg, T.~L. 2014.
\newblock ReferItGame: Referring to Objects in Photographs of Natural Scenes.
\newblock In \emph{Conference on Empirical Methods in Natural Language Processing}.

\bibitem[{Ke et~al.(2019)Ke, Pei, Li, Shen, and Tai}]{ke2019reflective}
Ke, L.; Pei, W.; Li, R.; Shen, X.; and Tai, Y.-W. 2019.
\newblock Reflective decoding network for image captioning.
\newblock In \emph{Proceedings of the IEEE/CVF international conference on computer vision}, 8888--8897.

\bibitem[{Kim et~al.(2024)Kim, Pertsch, Karamcheti, Xiao, Balakrishna et~al.}]{kim2024openvla}
Kim, M.~J.; Pertsch, K.; Karamcheti, S.; Xiao, T.; Balakrishna, A.; et~al. 2024.
\newblock OpenVLA: An Open-Source Vision-Language-Action Model.
\newblock arXiv:2406.09246.

\bibitem[{Krishna et~al.(2016)Krishna, Zhu, Groth, Johnson, Hata et~al.}]{krishna2016visual}
Krishna, R.; Zhu, Y.; Groth, O.; Johnson, J.; Hata, K.; et~al. 2016.
\newblock Visual Genome: Connecting Language and Vision Using Crowdsourced Dense Image Annotations.
\newblock arXiv:1602.07332.

\bibitem[{Laurençon et~al.(2023)Laurençon, Saulnier, Tronchon, Bekman, Singh et~al.}]{laurencon2023obelics}
Laurençon, H.; Saulnier, L.; Tronchon, L.; Bekman, S.; Singh, A.; et~al. 2023.
\newblock OBELICS: An Open Web-Scale Filtered Dataset of Interleaved Image-Text Documents.
\newblock arXiv:2306.16527.

\bibitem[{Li et~al.(2023{\natexlab{a}})Li, Li, Savarese, and Hoi}]{li2023blip2}
Li, J.; Li, D.; Savarese, S.; and Hoi, S. 2023{\natexlab{a}}.
\newblock BLIP-2: Bootstrapping Language-Image Pre-training with Frozen Image Encoders and Large Language Models.
\newblock arXiv:2301.12597.

\bibitem[{Li et~al.(2023{\natexlab{b}})Li, Du, Zhou, Wang, Zhao et~al.}]{li2023evaluating}
Li, Y.; Du, Y.; Zhou, K.; Wang, J.; Zhao, W.~X.; et~al. 2023{\natexlab{b}}.
\newblock Evaluating Object Hallucination in Large Vision-Language Models.
\newblock arXiv:2305.10355.

\bibitem[{Lin et~al.(2024)Lin, Tang, Ye, Cui, Zhu, Jin et~al.}]{lin2024moellava}
Lin, B.; Tang, Z.; Ye, Y.; Cui, J.; Zhu, B.; Jin, P.; et~al. 2024.
\newblock MoE-LLaVA: Mixture of Experts for Large Vision-Language Models.
\newblock arXiv:2401.15947.

\bibitem[{Liu, Emerson, and Collier(2023)}]{liu2023visual}
Liu, F.; Emerson, G.; and Collier, N. 2023.
\newblock Visual Spatial Reasoning.
\newblock arXiv:2205.00363.

\bibitem[{Liu et~al.(2023{\natexlab{a}})Liu, Lin, Li, Wang, Yacoob et~al.}]{liu2023lrv}
Liu, F.; Lin, K.; Li, L.; Wang, J.; Yacoob, Y.; et~al. 2023{\natexlab{a}}.
\newblock Mitigating Hallucination in Large Multi-Modal Models via Robust Instruction Tuning.
\newblock arXiv:2306.14565.

\bibitem[{Liu et~al.(2023{\natexlab{b}})Liu, Li, Li, and Lee}]{liu2023llava1.5}
Liu, H.; Li, C.; Li, Y.; and Lee, Y.~J. 2023{\natexlab{b}}.
\newblock Improved Baselines with Visual Instruction Tuning.
\newblock arXiv:2310.03744.

\bibitem[{Liu et~al.(2023{\natexlab{c}})Liu, Li, Wu, and Lee}]{liu2023llava}
Liu, H.; Li, C.; Wu, Q.; and Lee, Y.~J. 2023{\natexlab{c}}.
\newblock Visual Instruction Tuning.
\newblock arXiv:2304.08485.

\bibitem[{Loshchilov and Hutter(2019)}]{loshchilov2019adamw}
Loshchilov, I.; and Hutter, F. 2019.
\newblock Decoupled Weight Decay Regularization.
\newblock arXiv:1711.05101.

\bibitem[{Lu et~al.(2023)Lu, Schroecker, Gu, Parisotto, Foerster et~al.}]{lu2023structured}
Lu, C.; Schroecker, Y.; Gu, A.; Parisotto, E.; Foerster, J.; et~al. 2023.
\newblock Structured State Space Models for In-Context Reinforcement Learning.
\newblock arXiv:2303.03982.

\bibitem[{Lu et~al.(2022)Lu, Clark, Zellers, Mottaghi, and Kembhavi}]{lu2022unifiedio}
Lu, J.; Clark, C.; Zellers, R.; Mottaghi, R.; and Kembhavi, A. 2022.
\newblock Unified-IO: A Unified Model for Vision, Language, and Multi-Modal Tasks.
\newblock arXiv:2206.08916.

\bibitem[{Mercat et~al.(2024)Mercat, Vasiljevic, Keh, Arora, Dave, Gaidon, and Kollar}]{mercat2024linearizing}
Mercat, J.; Vasiljevic, I.; Keh, S.; Arora, K.; Dave, A.; Gaidon, A.; and Kollar, T. 2024.
\newblock Linearizing Large Language Models.
\newblock arXiv:2405.06640.

\bibitem[{OpenAI(2023)}]{openai2023gpt4v}
OpenAI. 2023.
\newblock GPT-4V(ision) System Card.

\bibitem[{Oquab et~al.(2024)Oquab, Darcet, Moutakanni, Vo, Szafraniec, Khalidov et~al.}]{oquab2024dinov2}
Oquab, M.; Darcet, T.; Moutakanni, T.; Vo, H.; Szafraniec, M.; Khalidov, V.; et~al. 2024.
\newblock DINOv2: Learning Robust Visual Features without Supervision.
\newblock arXiv:2304.07193.

\bibitem[{Penedo et~al.(2023)Penedo, Malartic, Hesslow, Cojocaru, Cappelli et~al.}]{refinedweb}
Penedo, G.; Malartic, Q.; Hesslow, D.; Cojocaru, R.; Cappelli, A.; et~al. 2023.
\newblock The {R}efined{W}eb dataset for {F}alcon {LLM}: outperforming curated corpora with web data, and web data only.
\newblock \emph{arXiv preprint arXiv:2306.01116}.

\bibitem[{Rafailov et~al.(2023)Rafailov, Sharma, Mitchell, Ermon, Manning, and Finn}]{rafailov2023direct}
Rafailov, R.; Sharma, A.; Mitchell, E.; Ermon, S.; Manning, C.~D.; and Finn, C. 2023.
\newblock Direct Preference Optimization: Your Language Model is Secretly a Reward Model.
\newblock arXiv:2305.18290.

\bibitem[{Share{GPT}(2023)}]{sharegpt}
Share{GPT}. 2023.
\newblock \url{https://sharegpt.com/}.

\bibitem[{Singh et~al.(2019)Singh, Natarajan, Shah, Jiang, Chen et~al.}]{singh2019vqa}
Singh, A.; Natarajan, V.; Shah, M.; Jiang, Y.; Chen, X.; et~al. 2019.
\newblock Towards VQA Models That Can Read.
\newblock arXiv:1904.08920.

\bibitem[{Smith, Warrington, and Linderman(2023)}]{smith2023s5}
Smith, J. T.~H.; Warrington, A.; and Linderman, S.~W. 2023.
\newblock Simplified State Space Layers for Sequence Modeling.
\newblock arXiv:2208.04933.

\bibitem[{Soboleva et~al.(2023)Soboleva, Al-Khateeb, Myers, Steeves, Hestness, and Dey}]{cerebras2023slimpajama}
Soboleva, D.; Al-Khateeb, F.; Myers, R.; Steeves, J.~R.; Hestness, J.; and Dey, N. 2023.
\newblock {SlimPajama: A 627B token cleaned and deduplicated version of RedPajama}.
\newblock \url{https://www.cerebras.net/blog/slimpajama-a-627b-token-cleaned-and-deduplicated-version-of-redpajama}.

\bibitem[{Taori et~al.(2023)Taori, Gulrajani, Zhang, Dubois, Li, Guestrin, Liang, and Hashimoto}]{taori2023alpaca}
Taori, R.; Gulrajani, I.; Zhang, T.; Dubois, Y.; Li, X.; Guestrin, C.; Liang, P.; and Hashimoto, T.~B. 2023.
\newblock Stanford Alpaca: An Instruction-following LLaMA model.
\newblock \url{https://github.com/tatsu-lab/stanford_alpaca}.

\bibitem[{Tong et~al.(2024)Tong, Liu, Zhai, Ma, LeCun, and Xie}]{tong2024eyes}
Tong, S.; Liu, Z.; Zhai, Y.; Ma, Y.; LeCun, Y.; and Xie, S. 2024.
\newblock Eyes Wide Shut? Exploring the Visual Shortcomings of Multimodal LLMs.
\newblock arXiv:2401.06209.

\bibitem[{Touvron et~al.(2023)Touvron, Lavril, Izacard, Martinet, Lachaux, Lacroix, Rozière, Goyal, Hambro, Azhar, Rodriguez, Joulin, Grave, and Lample}]{touvron2023llama}
Touvron, H.; Lavril, T.; Izacard, G.; Martinet, X.; Lachaux, M.-A.; Lacroix, T.; Rozière, B.; Goyal, N.; Hambro, E.; Azhar, F.; Rodriguez, A.; Joulin, A.; Grave, E.; and Lample, G. 2023.
\newblock LLaMA: Open and Efficient Foundation Language Models.
\newblock arXiv:2302.13971.

\bibitem[{Waleffe et~al.(2024)Waleffe, Byeon, Riach, Norick, Korthikanti et~al.}]{waleffe2024empirical}
Waleffe, R.; Byeon, W.; Riach, D.; Norick, B.; Korthikanti, V.; et~al. 2024.
\newblock An Empirical Study of Mamba-based Language Models.
\newblock arXiv:2406.07887.

\bibitem[{Wang et~al.(2023)Wang, Meng, Weng, He, Wu et~al.}]{wang2023lvis}
Wang, J.; Meng, L.; Weng, Z.; He, B.; Wu, Z.; et~al. 2023.
\newblock To See is to Believe: Prompting GPT-4V for Better Visual Instruction Tuning.
\newblock arXiv:2311.07574.

\bibitem[{Wightman(2019)}]{rw2019timm}
Wightman, R. 2019.
\newblock PyTorch Image Models.
\newblock \url{https://github.com/rwightman/pytorch-image-models}.

\bibitem[{Wu et~al.(2024)Wu, Fei, Qu, Ji, and Chua}]{wu2024nextgpt}
Wu, S.; Fei, H.; Qu, L.; Ji, W.; and Chua, T.-S. 2024.
\newblock NExT-GPT: Any-to-Any Multimodal LLM.
\newblock arXiv:2309.05519.

\bibitem[{Yan, Gu, and Rush(2023)}]{yan2023diffusion}
Yan, J.~N.; Gu, J.; and Rush, A.~M. 2023.
\newblock Diffusion Models Without Attention.
\newblock arXiv:2311.18257.

\bibitem[{Yu et~al.(2016)Yu, Poirson, Yang, Berg, and Berg}]{yu2016modeling}
Yu, L.; Poirson, P.; Yang, S.; Berg, A.~C.; and Berg, T.~L. 2016.
\newblock Modeling Context in Referring Expressions.
\newblock arXiv:1608.00272.

\bibitem[{Zhai et~al.(2023)Zhai, Mustafa, Kolesnikov, and Beyer}]{zhai2023sigmoid}
Zhai, X.; Mustafa, B.; Kolesnikov, A.; and Beyer, L. 2023.
\newblock Sigmoid Loss for Language Image Pre-Training.
\newblock arXiv:2303.15343.

\bibitem[{Zhang and Sennrich(2019)}]{zhang2019root}
Zhang, B.; and Sennrich, R. 2019.
\newblock Root Mean Square Layer Normalization.
\newblock arXiv:1910.07467.

\bibitem[{Zhang et~al.(2024)Zhang, Zeng, Wang, and Lu}]{zhang2024tinyllama}
Zhang, P.; Zeng, G.; Wang, T.; and Lu, W. 2024.
\newblock TinyLlama: An Open-Source Small Language Model.
\newblock arXiv:2401.02385.

\bibitem[{Zhao et~al.(2023)Zhao, Gu, Varma, Luo, Huang et~al.}]{zhao2023fsdp}
Zhao, Y.; Gu, A.; Varma, R.; Luo, L.; Huang, C.-C.; et~al. 2023.
\newblock PyTorch FSDP: Experiences on Scaling Fully Sharded Data Parallel.
\newblock arXiv:2304.11277.

\bibitem[{Zhou et~al.(2024)Zhou, Hu, Weng, Jia, Luo, Liu et~al.}]{zhou2024tinyllava}
Zhou, B.; Hu, Y.; Weng, X.; Jia, J.; Luo, J.; Liu, X.; et~al. 2024.
\newblock TinyLLaVA: A Framework of Small-scale Large Multimodal Models.
\newblock arXiv:2402.14289.

\bibitem[{Zhu et~al.(2023)Zhu, Chen, Shen, Li, and Elhoseiny}]{zhu2023minigpt4}
Zhu, D.; Chen, J.; Shen, X.; Li, X.; and Elhoseiny, M. 2023.
\newblock MiniGPT-4: Enhancing Vision-Language Understanding with Advanced Large Language Models.
\newblock arXiv:2304.10592.

\bibitem[{Zhu et~al.(2024)Zhu, Zhu, Liu, Ou, Mou, and Tang}]{zhu2024llavaphi}
Zhu, Y.; Zhu, M.; Liu, N.; Ou, Z.; Mou, X.; and Tang, J. 2024.
\newblock LLaVA-Phi: Efficient Multi-Modal Assistant with Small Language Model.
\newblock arXiv:2401.02330.

\end{thebibliography}

\section{Appendix}
The appendix is organized as follows:
\begin{itemize}
    \item We provide more detailed implementation details, including modality processing, model architecture, inference setup, and hardware systems in Section~\ref{sec:a1}.
    \item We demonstrate the significant impact of prompt order on Cobra, as mentioned in Section 4.1 of the main document, and present additional experiments in Section~\ref{sec:a2}.
    \item We present more examples of Cobra in terms of generation quality and its ability to overcome visual hallucinations in Section~\ref{sec:a3}.
\end{itemize}

\subsection{Implementation Details}
\label{sec:a1}

\noindent \textbf{Modality processing.} 
We utilize the default image transformations provided by torchvision and TIMM~\cite{rw2019timm} to implement all image processing operations. We naively resize all the images to the resolution of $384 \times 384$ and normalize pixel values according to the defaults defined by each pre-trained backbone, which often adhere to the traditional ImageNet defaults. We extract patch features from the penultimate layer, as done in other MLLM methods~\cite{liu2023llava}.

\noindent \textbf{Large language model (LLM).} The LLM backbone is initialized with the pre-trained weights from the Mamba chat model. We have chosen various open-source model weights, including Mamba models with 2.8 billion and 7 billion parameters, as the LLM backbone for our proposed model.

The Mamba-2.8B model\footnote{https://huggingface.co/state-spaces/mamba-2.8b-slimpj} was pre-trained on the SlimPajama dataset~\cite{cerebras2023slimpajama} consisting of 627 billion tokens, we also evaluate a model\footnote{https://huggingface.co/xiuyul/mamba-2.8b-zephyr} that underwent supervised fine-tuning on the UltraChat-200k dataset~\cite{ding2023enhancing}, as well as direct preference optimization~\cite{rafailov2023direct} on the UltraFeedback dataset~\cite{cui2023ultrafeedback}. 

The Mamba-7B model\footnote{https://huggingface.co/TRI-ML/mamba-7b-rw} is a base model, which was pre-trained on the RefinedWeb~\cite{refinedweb} dataset with 1.2T tokens and was not fine-tuned on any chat dataset.

\noindent \textbf{Prompt template.}
To maintain consistency with the instruction template of the pre-trained Mamba chat model, our prompt format follows the subsequent format: 

\begin{tcolorbox} 
    \raggedright
    \small
     \hspace{-6mm}

    $\texttt{<|user|>}$\\
    $X_{\texttt{instruct}}^1{\texttt{<|endoftext|>}}$\\
    $\texttt{<|assistant|>}$\\
    $X_{\texttt{answer}}^1{\texttt{<|endoftext|>}}$\\
    $\texttt{<|user|>}$\\
    $X_{\texttt{instruct}}^2{\texttt{<|endoftext|>}}$\\
    $\texttt{<|assistant|>}$\\

\end{tcolorbox}

For other base models that were not fine-tuned on a chat dataset, we use the following prompt template:
\begin{tcolorbox} 
    \raggedright
    \small
     \hspace{-6mm}

    $\texttt{In:}X_{\texttt{instruct}}^1$\\
    $\texttt{Out:}X_{\texttt{answer}}^1${\texttt{<|endoftext|>}}\\
    $\texttt{In:}X_{\texttt{instruct}}^2$\\
    $\texttt{Out:}$

\end{tcolorbox}

The text form of the prompt is processed by the same tokenizer that GPT-NeoX uses to obtain tokens, which are then passed through an embedding layer to obtain continuous embeddings. The embeddings obtained from passing the image through the encoder are directly concatenated to the beginning of the embedding sequence. This is then input into the Mamba model to start generating answers.

\noindent \textbf{Evaluation.} We fork the vlm-evaluation\footnote{https://github.com/TRI-ML/vlm-evaluation} tool as our evaluation tool on the benchmarks.

\noindent \textbf{Hardware.} For experiments with models of 2.8 billion parameters scale, the whole training process of a single model takes about 26.5 hours on 8 NVIDIA A100 80GB GPUs. During the training process, we use the PyTorch Fully Sharded Data Parallel~\cite{zhao2023fsdp} framework and enable automatic mixed-precision with FP32 and BF16 for distributed training. The batch size is set as 128. We employ the AdamW~\cite{loshchilov2019adamw} optimizer with a cosine decay learning rate to update the network parameters and set the learning rate to $2\times10^{-5}$, with a decay factor of 0.1 and a warm-up ratio of 0.03. The model is trained for 2 epochs via supervised fine-tuning. 

\begin{table}
\setlength{\tabcolsep}{1mm}
  \centering
  {
  \begin{tabular}{lcccc}
    \toprule
    Model & {OCR First} & {OCR Last} & {w/o OCR tokens}\\
    \midrule
    {LLaVA v1.5} & {-} & {58.2} & {46.1}\\
    \midrule
    \textbf{Cobra-3.5B} & {58.2} & {43.0} & {47.9}\\
    \textbf{w/ SigLIP} & {58.8} & {47.3} & {49.3}\\
    \textbf{w/ LDPv2} & {54.7} & {44.7} & {40.3}\\
    \textbf{w/ Base} & {57.9} & {47.6} & {47.9}\\
    \textbf{w/ 1 Ep FT} & {57.5} & {45.4} & {46.4}\\
    \textbf{w/ PT+FT} & {58.0} & {47.4} & {46.6}\\
    \textbf{Cobra-8B} & {59.5} & {43.0} & {50.7}\\
    \bottomrule
  \end{tabular}}
  \vspace{-1mm}
  \caption{Additional Results of TextVQA.}
  \label{tab:a1}
  \vspace{-2mm}
\end{table}
\subsection{Additional Evaluation on TextVQA}
\label{sec:a2}
Here, we present the experimental results of all our models on the TextVQA dataset, constructed according to the description of prompt order in Section 4.1. As in Table~\ref{tab:a1}, ``OCR First" represents placing the Reference OCR tokens before the question, while ``OCR Last" involves presenting the question first and then the Reference OCR tokens, which is also the default prompt format for LLaVA. It can be observed that all models perform significantly better when the OCR tokens are placed before the question, resulting in an accuracy improvement of over 10\%. In the ``OCR Last" mode, the performance of most models is even lower than that without OCR tokens as prompts (except "w/ LDPv2" and "w/ PT+FT"). We attribute this phenomenon to the inherent inductive bias of the RNN models.

\subsection{More Examples}
\label{sec:a3}
Here, we provide more examples generated by Cobra-3.5B and LLaVA v1.5-7B, which are shown in Table~\ref{tab:a2}--\ref{tab:a4}.

\begin{table*}[t]
    \centering 
    \scalebox{1.0}{
      \begin{tabular}{p{2cm}  p{14.5cm}}
        \toprule
        &  \includegraphics[height=3cm]{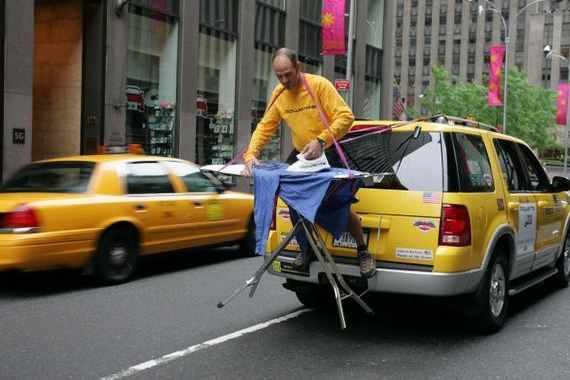} \\
        User & What is unusual about this image? \\
        \midrule
        Cobra (ours) & The unusual aspect of this image is that a man is ironing clothes \blue{while standing on the back of a yellow SUV parked on a city street}. This is not a typical scene, as people usually iron clothes indoors, in a designated ironing area, or at home. The man's choice to iron clothes on the back of a vehicle in a busy city street adds an element of surprise and humor to the image.\\
        \midrule
        LLaVA v1.5 & The unusual aspect of this image is that a man is ironing clothes while standing on the back of a moving car. This is not a typical scene, as ironing clothes is usually done indoors, in a stationary position, and not while the car is in motion. The man's ability to perform this task while standing on the back of a car is both unconventional and potentially dangerous, as it could lead to accidents or injuries. \\
        \bottomrule
      \end{tabular}}
    \caption{\textbf{Additional Example 1.} Cobra generates more detailed textual descriptions compared with the baseline, where \textcolor{blue}{blue} represents a more detailed description. }
    \label{tab:a2}  
    \vspace{-1mm}
\end{table*}

\begin{table*}[t]
    \centering 
    \scalebox{1.0}{
      \begin{tabular}{p{2cm}  p{14.5cm}}
        \toprule
        &  \includegraphics[height=3cm]{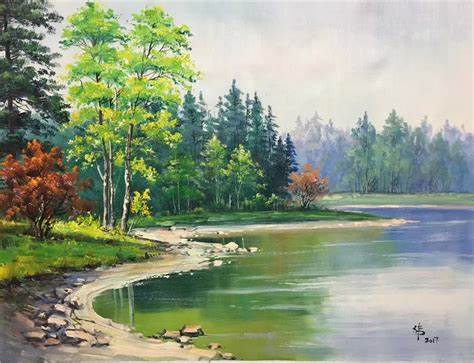} \\
        User & What's going on in this image? \\
        \midrule
        Cobra (ours) & The image showcases a serene landscape featuring a calm body of water surrounded by lush greenery. There are trees on the left and right sides of the water, \green{and a few rocks are scattered near the water's edge}. The sky above is clear, and the overall ambiance of the painting is tranquil and picturesque.\\
        \midrule
        LLaVA v1.5 & The image features a serene scene of a river surrounded by a lush forest. The river is calm, and the water appears to be green, giving it a tranquil atmosphere. The forest is filled with trees of various sizes, creating a picturesque landscape.

        \red{In the foreground, there is a bench situated near the water, inviting visitors to sit and enjoy the view. A bird can be seen perched on a branch}, adding to the natural beauty of the scene. The overall setting is perfect for relaxation and appreciation of nature.\\
        \bottomrule
      \end{tabular}}
    \caption{\textbf{Additional Example 2.} Cobra produced more accurate responses compared to the baseline, which suffered from severe hallucinations in describing the image, where \green{green} indicates a correct answer, \red{red} produces inaccurate answers.}
    \label{tab:a3}  
    \vspace{-1mm}
\end{table*}

\begin{table*}[t]
    \centering 
    \scalebox{1.0}{
      \begin{tabular}{p{2cm}  p{14.5cm}}
        \toprule
        &  \includegraphics[height=3cm]{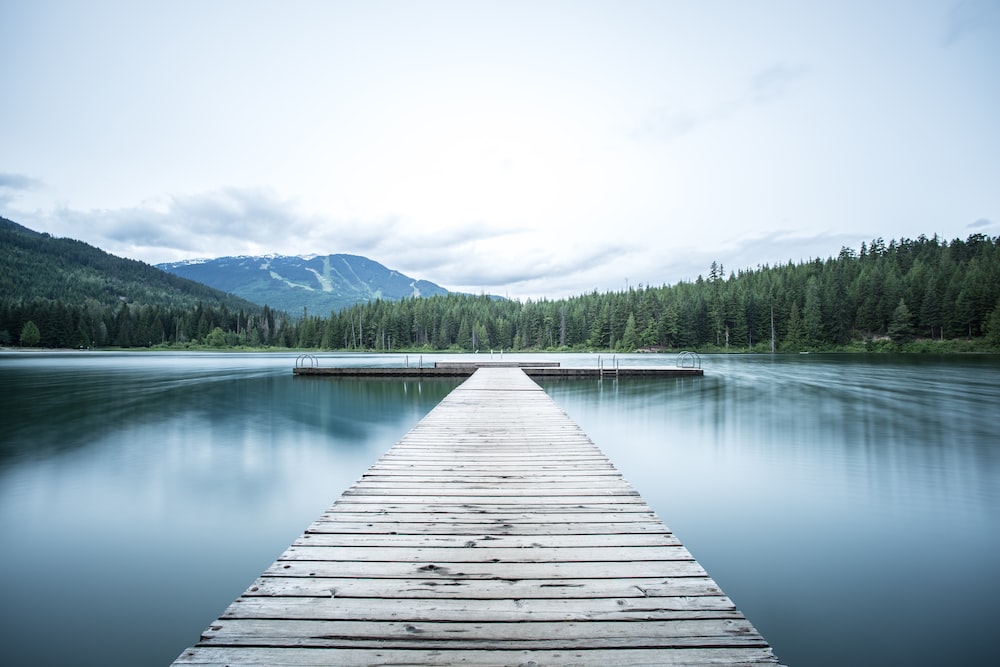} \\
        User & What are the things I should be cautious about when I visit here? \\
        \midrule
        Cobra (ours) & {When visiting this serene and picturesque location with a wooden dock extending over a calm lake, surrounded by a forest, and mountains in the background, you should be cautious about the following:

        1. Weather conditions: The image shows a cloudy day, which might indicate potential changes in weather, such as rain or storms. It's essential to be prepared for any sudden changes in weather conditions, such as bringing rain gear or checking the weather forecast before visiting.
        
        2. Water safety: The calmness of the lake might make it seem safe for swimming or other water activities, but it's important to be aware of the water currents, potential underwater hazards, and the presence of any wildlife.
        
        3. Wildlife: The forest surrounding the lake is home to various animals, such as birds, fish, and mammals. Be cautious not to disturb their natural habitat and respect their space.
        
        4. Navigation: If you plan to use the dock or any nearby boats, make sure you have the necessary skills and experience to safely operate them.
        
        5. Accessibility: The dock is located in a remote area, so it might be difficult to access in case of emergencies or if you need assistance. Make sure you have a reliable means of transportation to and from the location.
        
        By being cautious and prepared, you can enjoy a peaceful and safe visit to this beautiful location.} \\
        \midrule
        LLaVA v1.5 & When visiting the pier over the lake, there are a few things you should be cautious about. First, ensure that you have appropriate footwear with good traction, as the pier may be wet or slippery, especially if it has been raining or if there is algae on the surface. Second, be mindful of the weather conditions, as the image shows a cloudy day, which could indicate potential rain or storms. It is essential to stay safe and avoid being caught in a sudden downpour or storm. Lastly, be aware of your surroundings and the potential for wildlife, such as birds or fish, that might be present in the area. This will help you to enjoy your visit while minimizing any risks or disturbances to the local ecosystem. \\
        \bottomrule
      \end{tabular}}
    \caption{\textbf{Additional Example 3.} Compared with the baseline, Cobra produced responses that were clearer, more organized, and significantly more detailed.}
    \label{tab:a4}  
    \vspace{-1mm}
\end{table*}

\end{document}